\documentclass[10pt,twocolumn,letterpaper]{article}

\usepackage[pagenumbers]{cvpr}

\definecolor{cvprblue}{rgb}{0.21,0.49,0.74}
\usepackage[pagebackref,breaklinks,colorlinks,allcolors=cvprblue]{hyperref}

\newcommand{\topic}[1]{\smallskip\noindent\textbf{#1}}

\title{Move-in-2D: 2D-Conditioned Human Motion Generation}

\author{Hsin-Ping Huang$^{1,2}$  \,\, 
Yang Zhou$^1$  \,\, 
Jui-Hsien Wang$^1$  \,\, 
Difan Liu$^1$  \,\, 
\\
Feng Liu$^1$  \,\, 
Ming-Hsuan Yang$^2$  \,\, 
Zhan Xu$^1$
\\
[0.5em]
$^1$Adobe Research \,\, $^2$University of California, Merced
\\
[0.5em]
\normalsize
\url{https://hhsinping.github.io/Move-in-2D}
}

\begin{document}
\maketitle
\begin{abstract}
Generating realistic human videos remains a challenging task, with the most effective methods currently relying on a human motion sequence as a control signal. Existing approaches often use existing motion extracted from other videos, which restricts applications to specific motion types and global scene matching. We propose \textit{Move-in-2D}, a novel approach to generate human motion sequences conditioned on a scene image, allowing for diverse motion that adapts to different scenes. Our approach utilizes a diffusion model that accepts both a scene image and text prompt as inputs, producing a motion sequence tailored to the scene. To train this model, we collect a large-scale video dataset featuring single-human activities, annotating each video with the corresponding human motion as the target output. Experiments demonstrate that our method effectively predicts human motion that aligns with the scene image after projection. Furthermore, we show that the generated motion sequence improves human motion quality in video synthesis tasks.
\end{abstract}
    
\begin{figure*}[t]
    \centering
    \includegraphics[width=\textwidth]{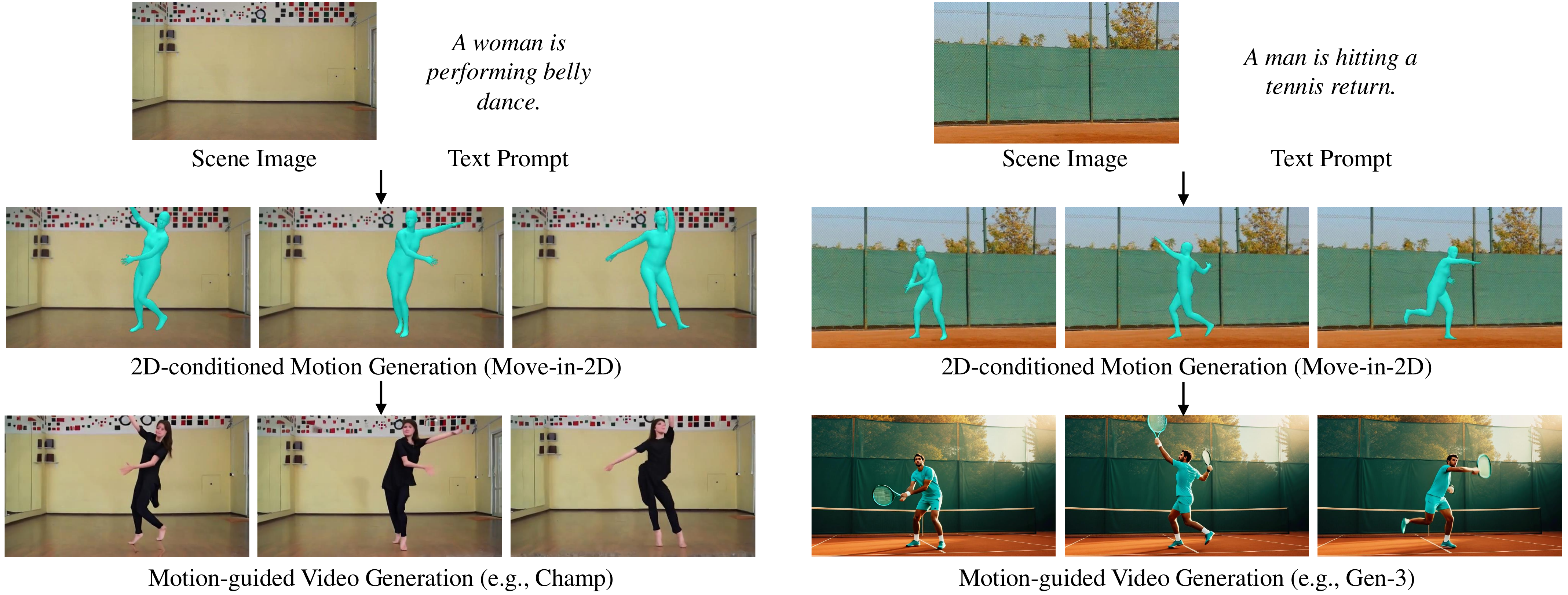}
    \vspace{-6mm}
    \caption{\textbf{2D-conditioned human motion generation.} Given an image representing the target scene and a text prompt describing the desired motion, we generate a motion sequence that aligns with the text description and projects naturally onto the scene image. 
This generated motion then serves as the control signal for the subsequent video generation tasks. 
}
    \label{fig:teaser}
    \vspace{-3mm}
\end{figure*}

\section{Introduction}
\label{sec:intro}
With the advancement of diffusion models, video generation has made significant progress. However, generating realistic human motion in a scene remains a nontrivial task due to the complexity of human movement. The human body is highly structured, and realistic motion generation requires models to learn and preserve articulations throughout the video. Many works~\cite{zhu2024champ,hu2023animateanyone,xu2023magicanimate,wang2023disco,dreampose_2023} have improved the quality of human videos by incorporating human-specific priors, specifically by adopting motion sequences as control signals during the generation process. These driving motion sequences are typically extracted from another video of the same class, with poses mostly aligned with the target human and minimal global motion. Consequently, although these approaches enhance the quality of generated human videos, they are still limited to specific motion domains (such as dancing) and no locomotion.

In this work, we propose to \textbf{generate} a motion sequence based on a 2D background rather than relying on pre-existing driving sequences. Formally, we define 2D-conditioned human motion generation as follows: given an image representing the target scene and a text prompt describing the desired motion, we generate a motion sequence that aligns with the text description and can be projected naturally onto the scene image. This approach enables a two-pass human video generation pipeline, as shown in~\cref{fig:teaser}. In the first pass, human poses are positioned using a template prior, preserving body articulation and generating a plausible motion sequence. This generated motion then serves as the control signal for the subsequent video generation. Compared to methods that rely on external motion sequences, 2D-conditioned motion generation can produce sequences that consistently align with the target background and text prompt description, without being constrained by specific motion types or minimal global movement.

Although human motion generation has been extensively studied, no existing approach directly addresses this novel setting. Some methods condition motion generation solely on text prompts~\cite{tevet2023human,chen2023executing,ahn2018text2action}, which, while straightforward, may not produce motion that seamlessly integrates into specific target environments, requiring further adjustments for scene compatibility. Other methods~\cite{wang2022humanise,huang2023diffusion} generate human motion based on 3D representations of the scene, such as 3D meshes or scanned point clouds. While these methods ensure scene affordance, obtaining 3D scenes is time-consuming and demands specialized equipment and manual effort. As a result, 3D scene-aware motion generation approaches are often limited to simple motion types (walk, sit, etc.) and indoor scenes.

Our 2D-conditioned approach introduces a new modality for human motion generation by incorporating affordance awareness through the input ~\textit{2D scene images}. This greatly expands the scope of existing approaches. A single 2D scene image provides semantic and spatial layout information about the target environment from a 2D perspective, enabling the generation of affordant human motion without the need for 3D scene reconstruction, especially for cases, e.g., video generation, when the motion is intended to be finally projected back onto a 2D plane. Furthermore, conditioning on 2D images allows for greater diversity in available scenes, as numerous online videos contain human activity in various environments. For example, outdoor scenes, which are hard to be used by 3D-aware motion generation networks, can be easily represented as 2D images and be consumed by the proposed approach.

On the other side, this novel setup also introduces several key challenges. First, training the model requires a dataset containing human motion sequences, text prompts describing the motion, and images representing the background scene. However, no existing dataset meets these requirements. Second, it remains unclear how to effectively condition the network on both text and scene image inputs. To address these challenges, we collect a large video dataset from internal data sources of open-domain internet videos.
We filter the videos to ensure a static background, so that any selected frame can reliably represent the scene throughout the motion sequence. We further annotate the human motion using a state-of-the-art 3D pose estimation approach~\cite{goel2023humans}. Leveraging this large-scale human motion dataset, we train a conditional diffusion model that generates human motion based on a single scene image and a text prompt. Inspired by in-context learning in large language models (LLMs)~\cite{an-etal-2023-context,wei2022emergent,Peebles2022DiT}, we employ a similar strategy to convert scene and text inputs into a shared token space, integrating them within a transformer-based diffusion model for the output.

Our contributions can be summarized as follows:

\begin{itemize}

\item We introduce a novel task of generating human motion with a 2D image and text as conditions.
It provides a more accessible way to motion generation by incorporating scene conditions without requiring 3D reconstruction.

\item We collect a large human video dataset with annotated 3D human motion.
The dataset significantly increases the scale of existing scene-aware motion generation datasets.

\item We propose a diffusion-based network conditioned on both text and input scene images. %
We also show that the output motion is able to improve the quality of human motion when generating videos.

\end{itemize}

\section{Related Work}
\topic{Human video generation.}
Human-centric video generation typically leverages ControlNet to condition the generation process on motion guidance signals, such as OpenPose~\cite{wang2023disco,hu2023animateanyone}, DensePose~\cite{xu2023magicanimate,dreampose_2023} keypoints, or SMPL mesh sequences~\cite{zhu2024champ}. %
While these approaches deliver visually plausible results, they depend on predefined motion sequences as guidance, thus restricting their ability to generate diverse motions.
In contrast, we address an orthogonal problem: generating motion guidance sequences conditioned on text prompts and a scene image. Our generated motion can then be used as guidance within human video generation frameworks. %

\topic{Human motion datasets.}
Several datasets have been proposed to facilitate research on human motion understanding and generation. Datasets such as CMU Mocap~\cite{cmu_mocap}, Human3.6M~\cite{h36m_pami}, and MoVi~\cite{ghorbani2021movi} capture human motion but lack textual descriptions of the actions. The KIT Motion Language Dataset~\cite{Plappert2016} provides both motion sequences and textual prompts, containing approximately 3.9K motion sequences. HumanML3D~\cite{Guo_2022_CVPR} expands this number to 14.6K by sourcing motion data from HumanAct12~\cite{guo2020action2motion} and AMASS~\cite{AMASS:ICCV:2019}. %
The Motion-X dataset~\cite{lin2023motionx} scales up further to 81K motion sequences, and includes not only body movement but also facial expressions and hand poses. Despite the increasing scale of these datasets, none provide scene context aligned with the motion sequences.

Some datasets provide captured 3D scenes alongside motion sequences. Works such as \cite{ren2023lidar, yan2023cimi4d} include SMPL models with global positions in 3D scenes. The PROX dataset~\cite{PROX} utilizes optimization techniques with RGB-D data to reconstruct human motions, while others \cite{araujo2023circle, wang2021synthesizing} compile synthesized data of human-scene interactions. Additionally, datasets like \cite{wang2022humanise, zhao2022compositional} incorporate both scene context and language descriptions for specific actions. However, these datasets primarily focus on indoor environments due to the challenges of 3D scene representation. Furthermore, some are oriented toward global motion prediction, resulting in limited scene diversity and a lack of detailed textual annotations.

\topic{Text-driven human motion generation.} With the availability of motion datasets, human motion generation has made notable progress. Early approaches, such as Text2Action~\cite{ahn2018text2action}, utilized recurrent neural networks to capture temporal dependencies within motion sequences. Later, transformer-based architectures were introduced in works like TM2T~\cite{guo2022tm2t} and TEACH~\cite{athanasiou2022teach}, enabling improved control and the generation of longer, more coherent sequences. Building on these advancements, models such as MotionGPT~\cite{jiang2024motiongpt} leverage large language model (LLM) pretraining for text-driven motion synthesis.

More recently, many works have applied diffusion models in motion generation tasks. MotionDiffuse~\cite{zhang2022motiondiffuse} and MDM~\cite{tevet2023human} generate motion sequences aligned with text prompts from an initial random noise. ReMoDiffuse~\cite{zhang2023remodiffuse} introduces a retrieval-augmented model, where knowledge from retrieved samples enhances motion synthesis. MLD~\cite{chen2023executing} performs motion diffusion in a latent space using a variational autoencoder. EMDM~\cite{zhou2025emdm} further reduces the number of sampling steps required during the denoising process. %
Although motion can be generated with only a text prompt, these methods often lack contextual alignment with specific virtual environments, limiting their direct applicability as control signals in video generation.

\topic{Scene-aware human motion generation.} Given a 3D indoor scene, prior works~\cite{zhang2020generating, zhang2020generating, hassan2021populating, hassan2021stochastic, wang2021synthesizing, wang2021scene} have demonstrated the generation of physically plausible human poses or motion sequences. HUMANISE~\cite{wang2022humanise} aligns captured human motion sequences with various 3D indoor scenes, using text prompts as conditioning inputs. LaserHuman~\cite{cong2024laserhuman} incorporates real human motions within 3D environments, supporting open-form descriptions, and spanning both indoor and outdoor settings. However, due to the challenges associated with obtaining 3D scenes, these methods are often trained on limited datasets, which constrains their generalizability to more diverse, in-the-wild backgrounds.

\label{sec:related}

\begin{figure*}[t]
    \centering
    \includegraphics[width=0.8\textwidth]{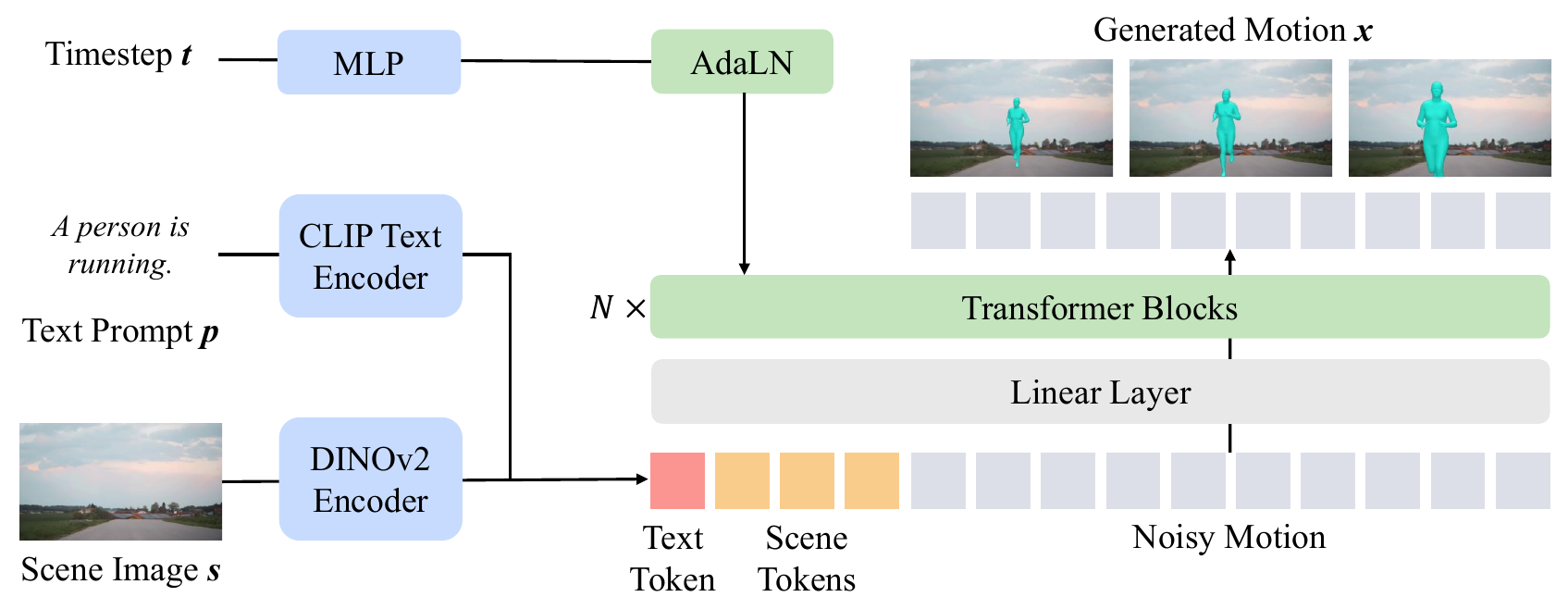}
    \vspace{-2mm}
    \caption{\textbf{Overview.} The text prompt and background scene image are encoded by the CLIP and DINO encoders, and incorporated into the model via in-context conditioning. The AdaLN layer receives the diffusion timestep as input. Our multi-conditional transformer model then generates a human motion sequence through a diffusion denoising process, aligning the generated motion with both input conditions.}%
    \label{fig:method}
     \vspace{-3mm}
\end{figure*}

\begin{table}[t]
    \centering
    \scriptsize
    \caption{
\textbf{Dataset statistics.} HiC-Motion is the largest dataset comprising motions, text, and diverse indoor and outdoor scenes.}
\vspace{-2mm}
        \tabcolsep=0.05cm
    \begin{tabular}{lccc|c|c}
        \toprule
        Dataset & Motions & Texts & Scenes & Scene Representation & Scene Type\\
        \midrule
        KIT~\cite{Plappert2016} & 3.9k & 6.2k & No & No & Indoor \\
        HumanML3D~\cite{Guo_2022_CVPR} & 14.6k & 44.9k & No & No & Indoor \\
        HUMANISE~\cite{wang2022humanise} & 19.6k & 19.6k & 643 & RGBD & Indoor \\
        PROX~\cite{PROX} & 28k & No & 12 & RGBD & Indoor \\
        LaserHuman~\cite{cong2024laserhuman} & 3.5k & 12.3k & 11 & RGBD & Indoor/Outdoor\\
        Motion-X~\cite{lin2023motionx} & 81.1k & 81.1k & 81.1k & Video & Indoor/Outdoor\\
        \midrule
        HiC-Motion & 300k & 300k & 300k & Video & Indoor/Outdoor \\
        \bottomrule
    \end{tabular}
    \label{tab:statistics}
    \vspace{-2mm}

\end{table}

\section{Humans-in-Context Motion Dataset}
\label{sec:dataset}
To advance human motion generation in 2D scenes, a large-scale video dataset capturing open-domain human motions in diverse scenes is crucial.
Human-centric video datasets, such as HiC~\cite{kulal2023affordance,brooks2021hallucinating,pan2024actanywheresubjectawarevideobackground}, provide millions of video clips but lack motion and text annotations. Additionally, these datasets are limited by short sequence lengths and low spatial resolutions. See~\cref{sec:related} and~\cref{tab:statistics} for more discussions.
Inspired by HiC, we collect \textbf{Humans-in-Context Motion (HiC-Motion)}, a large-scale dataset of human motions capturing rich background scenes and natural language captions. 
Next, we describe our data collection and preprocessing pipeline.

\topic{Data collection.}
While action recognition datasets~\cite{Kay2017TheKH,monfortmoments} inherently include human actions, they are generally limited to short sequences and close-up shots, often omitting full-body views and background context.
Our dataset is sourced from an internal dataset containing 30M open-domain internet videos.
Despite the large pool, a significant portion of these videos lack human subjects.
We filter for videos containing a single human moving within the scene using keypoint-based models, including Keypoint R-CNN for person detection~\cite{Detectron2018} and OpenPose for keypoint prediction~\cite{8765346}, following~\cite{brooks2021hallucinating}. We retain videos with motion sequences exceeding 256 frames, resulting in a curated set of 300k videos—approximately 1\% of the initial dataset.
Our dataset includes high-quality, real-world videos with a range of indoor and outdoor scenes and diverse human activities, spanning daily tasks (e.g., drinking coffee, using a laptop) and sports (e.g., playing tennis, performing lunges), across more than 1k categories.

\topic{Data preprocessing.} To obtain human motion annotations from the selected videos, we use the off-the-shelf method 4D-Humans~\cite{goel2023humans} to extract pseudo ground-truth motions in SMPL format, ensuring high quality and frame-to-frame consistency.
Since our objective is to condition human motion on scene backgrounds, we utilize Mask R-CNN to detect person masks and apply a basic inpainting model~\cite{itseez2015opencv} to remove humans from the video frames. During training, we randomly select an inpainted frame from each video to serve as the background image.
To enhance the model’s generalization to unseen scenes, we apply color adjustments to simulate diverse lighting conditions in the background images~\cite{Karras2020ada} as well as random cutout augmentations.

\section{Approach}
\label{sec:approach}
Given a text prompt and a background scene image, our objective is to generate a human motion sequence that aligns with the action described in the text prompt while maintaining physical compatibility with the background scene.
We begin with a preliminary overview of diffusion model in \cref{sec:diffusion}. In \cref{sec:mdm}, we propose a conditional motion diffusion model. We then introduce a multi-conditional transformer in \cref{sec:dit}. Finally, in \cref{sec:training}, we present our training strategy. \cref{fig:method} provides an overview of our approach.

\subsection{Preliminaries on Diffusion Models}
\label{sec:diffusion}
Diffusion models like DDPM~\cite{ho2020denoising,tevet2023human} approximate the data distribution via a forward and backward process. In the forward process, Gaussian noise is added to the sample \(\mathbf{x}_0\), resulting in \(\mathbf{x}_t\). The model \(\mathcal{M}\) learns to reverse this process by denoising \(\mathbf{x}_t\) conditioned on timestep \(t\) and context \(c\). The training minimizes the MSE loss between the predicted clean sample \(\hat{\mathbf{x}}_0 = \mathcal{M}(\mathbf{x}_t | t, c)\) and the ground truth \(\mathbf{x}_0\), i.e., $\mathcal{L}_{\text{mse}} = \mathbb{E}_{\mathbf{x}_0, t} \left\| \mathbf{x}_0 - \mathcal{M}(\mathbf{x}_t | t, c) \right\|^2$.
During sampling, the model iteratively predicts \(\hat{\mathbf{x}}_0\) at each timestep \(t\) over \(T\) steps to recover \(\mathbf{x}_0\). 
Classifier-free guidance (CFG)~\cite{Ho2022ClassifierFreeDG} is applied to enhance alignment with conditions.

\subsection{Conditional Motion Diffusion}
\label{sec:mdm}
Given input conditions, including a text prompt \( p \) and a background scene image \( s \), we train a conditional motion diffusion model to generate the target human motion \( \mathbf{x} \).
The target human motion is represented as a sequence of \( N \) human poses, each with a dimensionality of \( D \). Each pose is parameterized by body pose parameters \( \theta_{b} \in \mathbb{R}^{23 \times 6} \), which capture 6D rotations for 23 SMPL joints~\cite{SMPL}, along with a global orientation parameter \( \theta_{g} \in \mathbb{R}^{6} \) that defines the overall human body orientation.
Unlike previous motion generation approaches~\cite{tevet2023human,chen2023executing}, we aim to generate human motion that projects naturally onto a 2D background scene image. Therefore, our model predicts an additional camera translation parameter \( \pi \in \mathbb{R}^{3} \), assuming a perspective camera with fixed focal length and intrinsics to project SMPL space points onto the image plane.
We train the conditional diffusion model \( \mathcal{M} \) by randomly dropping the text prompt \( p \) and background scene \( s \) conditions by a probability of $q$ where $q=0.1$ in our experiments. %
During sampling, we apply CFG to both the text and scene conditions jointly to enhance the alignment between the generated motion and the input conditions, where \(g\) is the guidance scale.
\begin{align}   
    \mathcal{M}_{\text{cfg}} = \mathcal{M}(\mathbf{x}_t| t) + g \left( \mathcal{M}(\mathbf{x}_t | t, p, s) - \mathcal{M}(\mathbf{x}_t | t) \right).
\end{align}

\subsection{Multi-Conditional Transformer}
\label{sec:dit}
We inject the text prompt and scene conditions into a diffusion transformer model to generate a motion sequence that aligns semantically with the input description and is physically compatible with the scene after projection to 2D.
Our model architecture follows the diffusion transformer~\cite{Peebles2022DiT,tevet2023human}.
The motion sequence \(\mathbf{x} \in \mathbb{R}^{T \times D} \) is projected into the transformer's hidden size, with positional embeddings added to the tokens before feeding them into a series of transformer blocks. The output tokens are then linearly projected back to obtain the motion prediction \( \hat{\mathbf{x}} \).

We now describe the condition encoding and injection process. We encode the diffusion timestep \( t \) by a positional embedding layer, the text prompt \( p \) by a CLIP encoder~\cite{Radford2021LearningTV}, and the background image condition \( s \) by a DINO encoder~\cite{oquab2023dinov2} which preserves the spatial relationships across patches. Each condition is then projected to the transformer dimension.
To guide the diffusion process, we employ simple yet effective methods~\cite{Peebles2022DiT} for injecting the conditions into the transformer blocks:
\begin{itemize}
    \item \textbf{In-context conditioning.} The conditions are concatenated to the motion sequence as additional tokens, which are removed from the output sequence without being calculated the loss.
    \item \textbf{Adaptive layer normalization (AdaLN)~\cite{perez2018film}.} 
    A linear layer predicts the scale and shift parameters from the condition tokens, which are then applied to the motion sequence.
    \item \textbf{Cross-attention layer.} A cross-attention layer is inserted after the self-attention layer to take the input conditions.
\end{itemize}
We observe that in-context learning for both the text and scene modalities improves the model's ability to capture interactions between the inputs by converting them into a shared token space, leading to better alignment with both conditions. Using AdaLN for the diffusion timestep condition enhances the temporal smoothness of the generated motions. We thus adopt this configuration as our main framework (see~\cref{sec:ablation}).

\subsection{Training Strategy}
\label{sec:training}
We adopt a two-stage training strategy to generate human motion aligned with text prompts and backgrounds. It first learns to generate diverse motion sequences and then to disentangle human motion from camera effects.

\topic{Selection of fine-tuning set.}
Our motion sequences are extracted from internet videos, which include both human and camera motion. For example, camera movement to the right can cause the pose sequence to shift left. 
However, our goal is to represent the scene using a single image, independent of any camera motion.
To achieve this, we calculate the optical flow of the raw videos and use the median flow value to select videos with minimal background motion. Additionally, to address data distribution biases where many daily activities involve limited human movement (e.g., sitting), we select a subset of videos with significant human motion (exceeding 200 pixels of movement) to encourage the generation of sequences with large motion dynamics.

\topic{Two-stage training.}
We adopt a two-stage training strategy to generate human motion aligned with text prompts and backgrounds. In the first stage, the model is trained on the full 300k video dataset for 600k iterations to learn scene semantics and generate diverse motion sequences based on text prompts. 
In the second stage, we fine-tune the model on a mixed dataset of 150k videos, consisting of 60\% large-motion videos and 40\% fixed-background videos, for an additional 600k iterations. This fine-tuning enhances the model's ability to generate human motion that decouples camera-induced movement and improves the generation of large-motion dynamics.

\begin{figure*}[t]
    \centering
    \includegraphics[width=\textwidth]{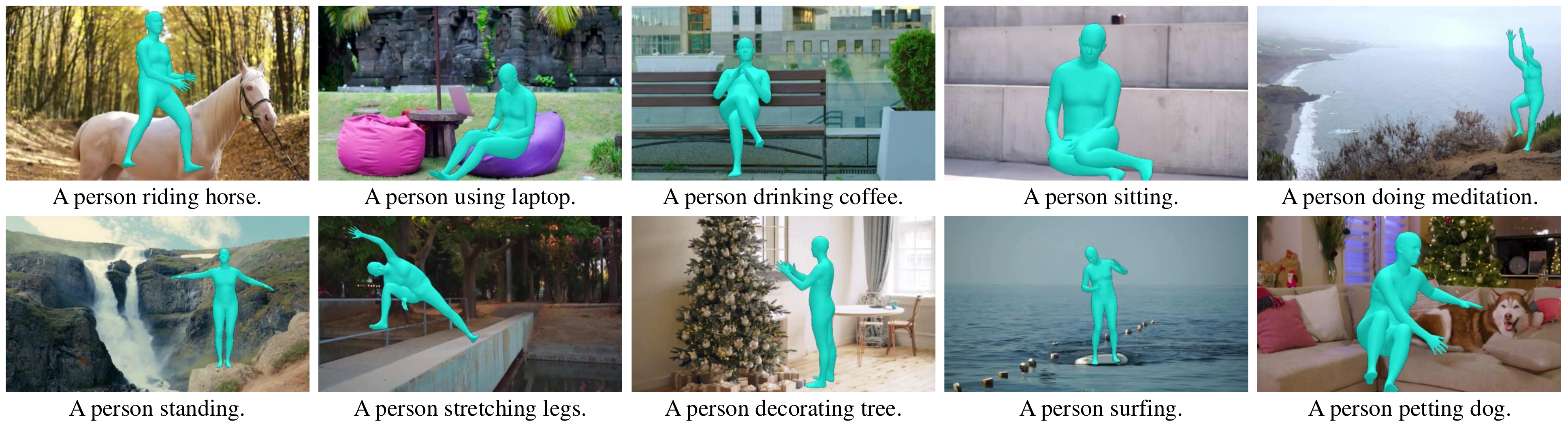}
    \vspace{-4mm}
    \caption{\textbf{Affordance-aware human generation.} Our model generates human poses consistent with both text prompts and scene context, such as standing on a cliff. It also supports complex human-scene interactions, including activities like petting a dog.
}
    \label{fig:static}
    \vspace{-2mm}
\end{figure*}

\begin{figure*}[t]
    \centering
    \includegraphics[width=\textwidth]{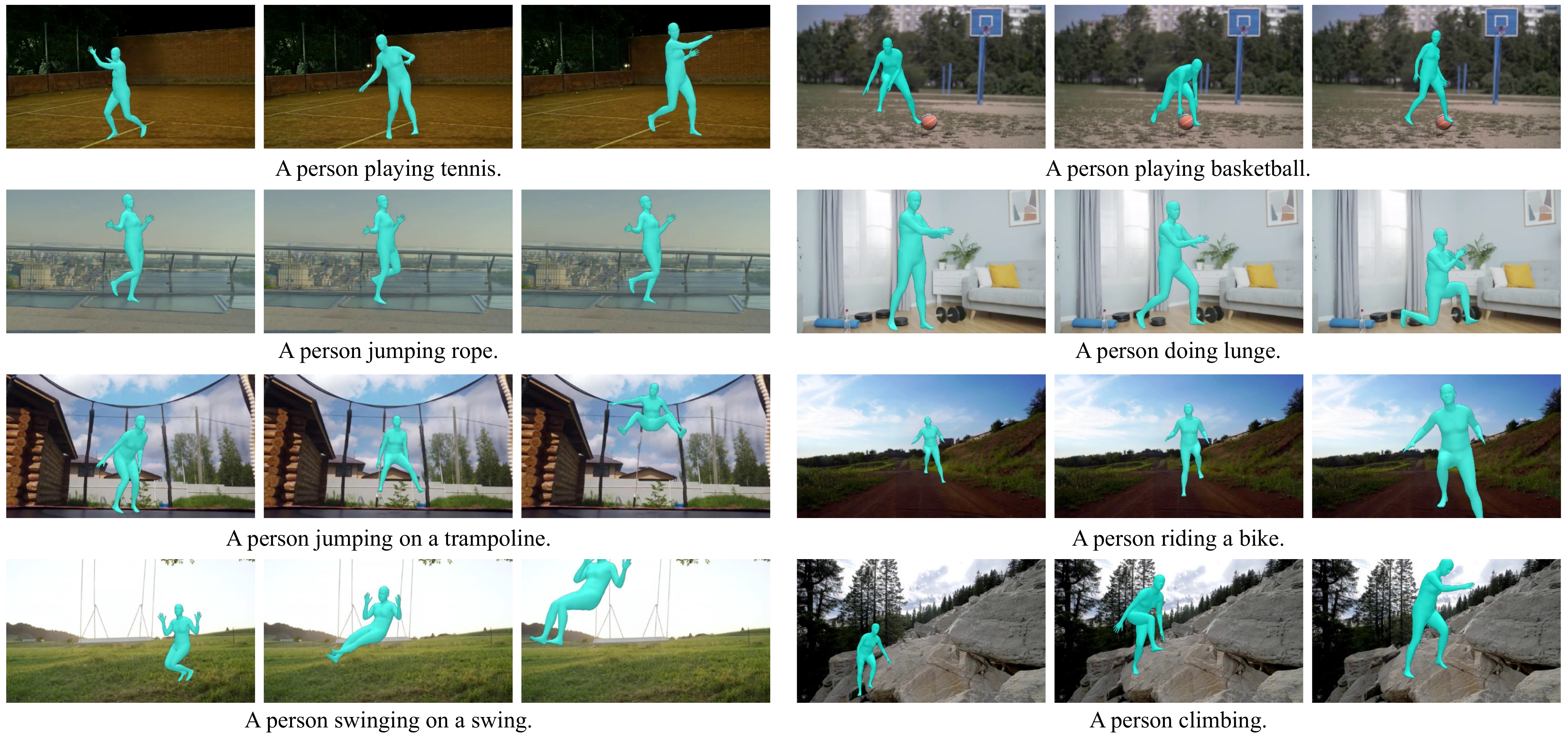}
    \vspace{-4mm}
    \caption{\textbf{Motion generation with large dynamics.} Our results show motion sequences that are accurately placed and move within scenes, such as playing tennis, enabling the generation of complex human activities that are challenging for video generation models.
    }
    \label{fig:large}
    \vspace{-2mm}
\end{figure*}

\section{Experiments}
\label{sec:exp}

\topic{Evaluation data.} To address the lack of benchmarks for evaluating human motion generation conditioned on 2D scenes, we construct a test set comprising text prompts, scene images, and ground-truth motion sequences for comprehensive evaluation.
Our test set is sampled from a held-out portion of the HiC-Motion dataset. We first curate 100 high-frequency verb phrases from the data to serve as text prompts (e.g., ``A person drinks coffee'').
For each text prompt, we sample 10 videos and randomly select one frame, where the human is removed, as the corresponding scene image. This process results in a total of 957 test samples.

\topic{Evaluated methods.}  
To assess the effectiveness of the proposed models in generating human motions that align with text prompts and are compatible with scene images, we compare them against state-of-the-art motion generation models conditioned on single or multiple modalities.
Specifically, \textit{MDM}~\cite{tevet2023human} and \textit{MLD}~\cite{chen2023executing} generate motion conditioned solely on text prompts.
While no existing methods generate motion conditioned on 2D scene images, we include models that utilize 3D point clouds to produce affordance-aware motion. We first employ a pretrained depth prediction model~\cite{depth_anything_v2} to estimate scene depth, then back-project our 2D scene images into 3D point clouds as the input conditions to baselines.
We compare to the following scene-conditioned approaches: \textit{SceneDiff}~\cite{huang2023diffusion}, which uses 3D point clouds as input, and \textit{HUMANISE}~\cite{wang2022humanise}, the closest approach to ours, which conditions on both text prompts and 3D point clouds.
Additionally, we extend MDM by training it on our HiC-Motion dataset, denoted as \textit{MDM+}.
We also evaluate two variants of our model:
\textit{Ours}, conditioned on both text prompts and scene images.
\textit{Ours-scene}, conditioned solely on scene images.

\topic{Evaluation metrics.}
To evaluate the quality and diversity of the generated human motions, previous works~\cite{guo2020action2motion, petrovich21actor} utilize a pre-trained human motion classifier to extract motion features for evaluation. However, due to the lack of motion feature extractors trained on open-domain videos, we train our classifier based on STGCN~\cite{petrovich21actor,stgcn} using motion sequences of length 256, with each pose represented by 21 SMPL joints in 6D rotation format. To standardize outputs across models, we ignore global orientation and translation.
We evaluate the models using four metrics:
\begin{itemize} 
\item \textbf{FID} assesses the overall quality of the generated motions by computing the distance between the feature distributions of generated and real motions. 
\item \textbf{Accuracy} evaluates the alignment between generated motions and input prompts by calculating the recognition accuracy of the generated motions. %
\item \textbf{Diversity} quantifies the variation across generated motions by calculating the distance between two randomly sampled subsets of generated motions from all prompts. 
\item \textbf{Multimodality} measures variation within identical prompts by computing the distance between two subsets of motions generated from the same prompts. 
\end{itemize}

\begin{figure*}[t]
    \centering
    \includegraphics[width=\textwidth]{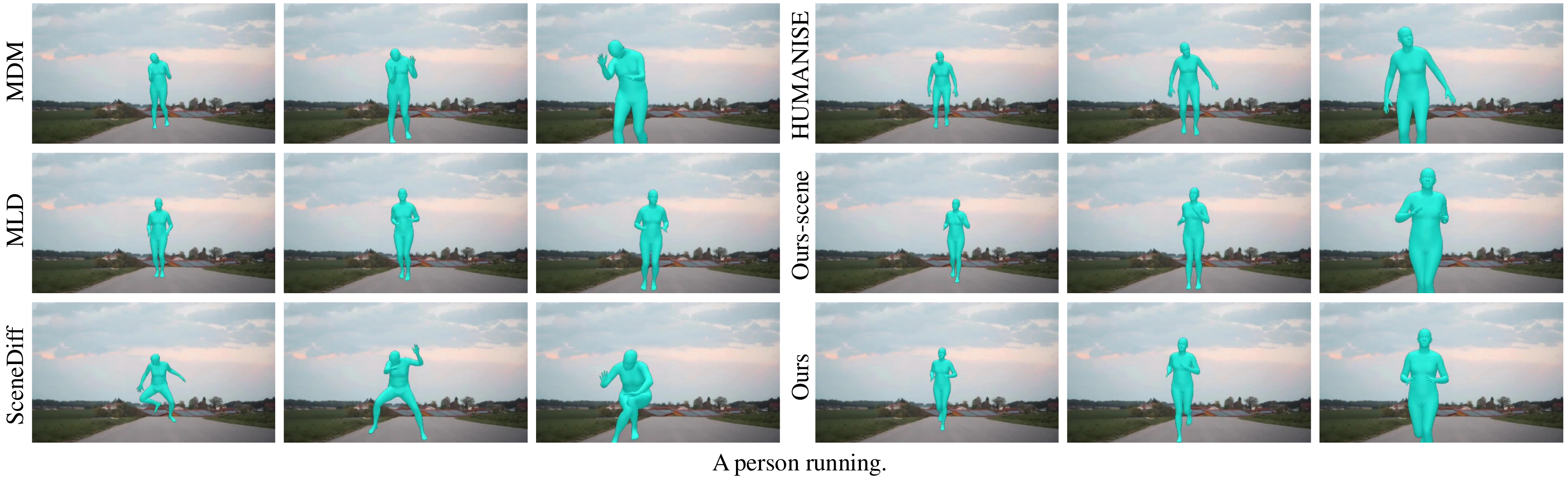}
    \vspace{-6mm}
    \caption{\textbf{Comparison to state-of-the-art.} \textit{MDM} and \textit{SceneDiff} produces implausible poses, \textit{MLD} generates mismatched motion with the scene, and \textit{HUMANISE} generates static poses. 
    Our method generates coherent motion aligned with both the scene and text prompts.}
    \label{fig:comparison}
    \vspace{-2mm}
\end{figure*}

\topic{Implementation details.}
Our model generates motion sequences of length \( N=256 \) and feature dimension \( D=147 \), using an architecture with 8 transformer blocks, 512 hidden units, feedforward layers of size 2048, and 4 attention heads. It is trained with the Adam optimizer with a learning rate of 0.0002, batch size 128 for 1.2M iterations, and 1000 diffusion steps with a cosine noise schedule. Scene images with resolution \( 168 \times 280 \) are encoded into 240 tokens by DINO-B~\cite{oquab2023dinov2}, and text prompts into a single token using CLIP-B~\cite{Radford2021LearningTV}, resulting in a sequence of 497 tokens.

\subsection{Qualitative Results}
\topic{Affordance-aware human generation.} 
\cref{fig:static} shows that our model generates human poses that are consistent with both the text prompts and the scene context, such as standing at the edge of a cliff, sitting on a chair, and surfing on a board. 
In addition, our model is capable of generating complex human-scene interactions, including activities such as riding a horse, decorating a tree, and petting a dog.

\topic{Motion generation with large dynamics.}
In \cref{fig:large}, we present examples with larger motion dynamics. Our results show strong scene compatibility, with the human sequence correctly placed and moving in environments like jumping on a trampoline.
Our model generates complex human activities with detailed pose sequences aligned with text prompts, such as playing tennis, which is challenging for video generation models and is effectively handled by our approach.

\topic{Comparison to state-of-the-art.}
As shown in~\cref{fig:comparison}, the text-conditioned method \textit{MDM} fails to generate plausible poses. \textit{MLD} correctly generates the running action, but the motion is not compatible with the scene, especially since it fails to generate the person moving toward the camera.
The scene-conditioned method \textit{SceneDiff} struggles to generate accurate human poses, while \textit{HUMANISE} produces static poses throughout the sequence. These methods, trained on limited synthetic point cloud data, have difficulty adapting to real-world scene conditions.
In contrast, our method generates motion that is both coherent within the scene and aligned with the text prompts.

\begin{table}[t]
    \centering
    \scriptsize
    \caption{\textbf{Quantitative results.} Our method achieves better quality and diversity scores compared to state-of-the-art text-conditioned, scene-conditioned, and multimodal motion generation models.}
        \tabcolsep=0.1cm
         \vspace{-1mm}
    \begin{tabular}{lcccc}
        \toprule
        Methods & FID~$(\downarrow)$ & Accuracy~$(\uparrow)$ & Diversity~$(\uparrow)$ & Multimodality~$(\uparrow)$\\
        \midrule
        MDM~\cite{tevet2023human} & 164.595 & 0.325 & 24.758 & 18.924 \\
        MLD~\cite{chen2023executing} & 85.913 & 0.322 & 25.119 & 19.464 \\
        \midrule
        SceneDiff~\cite{huang2023diffusion} & 543.769 & 0.203 & 4.217 & 3.861\\
        HUMANISE~\cite{wang2022humanise} & 159.935 & 0.225 & 23.287 & 19.956 \\
        \midrule
        MDM+~\cite{tevet2023human} & 46.035 & 0.620 & 23.002 & 17.627 \\
        Ours-scene & 46.458 & 0.482 & 24.968 & \bf 21.320 \\
        Ours & \bf 44.639 & \bf 0.661 & \bf 26.027 & 20.130 \\
        \bottomrule
    \end{tabular}
    \label{tab:quantitative}

\end{table}

\begin{figure*}[t]
    \centering
    \includegraphics[width=\textwidth]{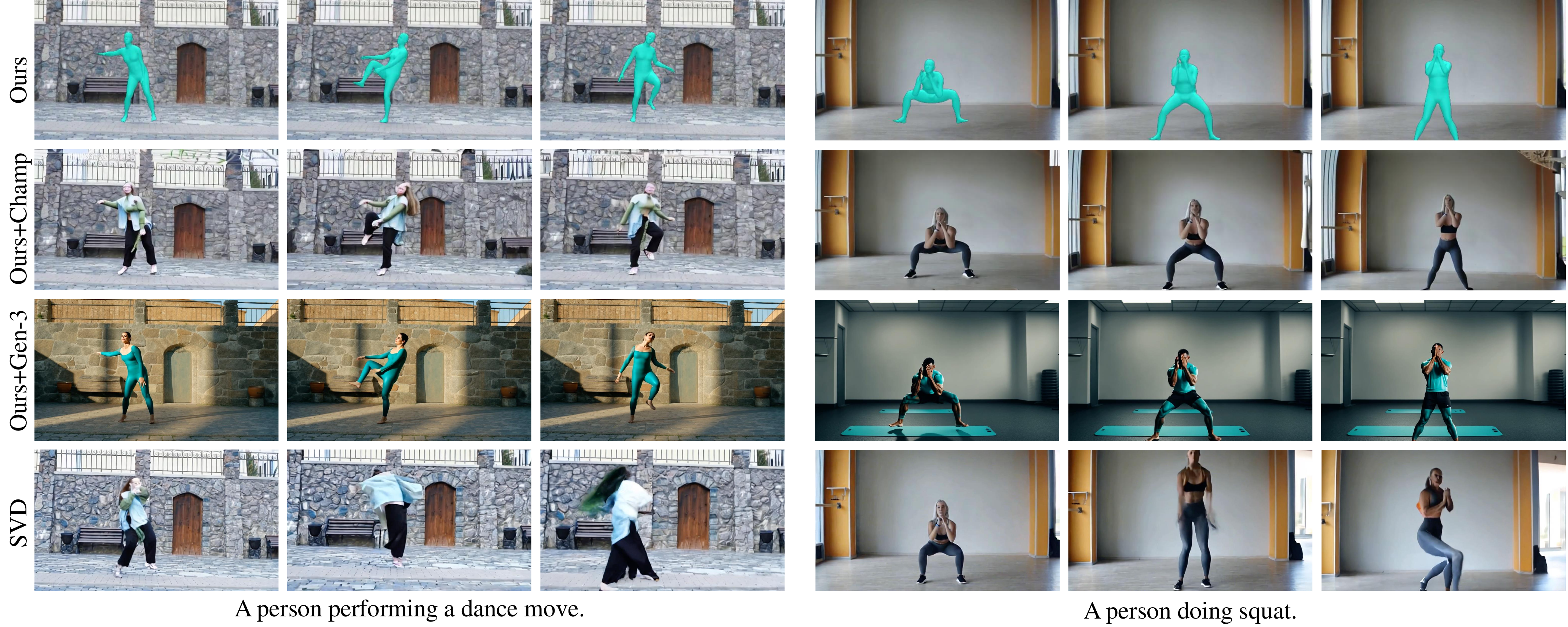}
    \vspace{-5.5mm}
    \caption{\textbf{Motion-guided human video generation.} 
Our approach generates scene-compatible motion sequences from a scene image and text prompt, which are then used to animate a reference human using Champ~\cite{zhu2024champ} or Gen-3~\cite{gen1}. The generated motion ensures accurate human shapes and smooth motion in the resulting videos, outperforming SVD~\cite{blattmann2023stablevideodiffusionscaling} in preserving human geometry and motion consistency.    
}
    \vspace{-2mm}
    \label{fig:champ}
\end{figure*}

\subsection{Quantitative Results}

The evaluation results are shown in~\cref{tab:quantitative}. We observe that the scene-conditioned motion generation models, \textit{HUMANISE} and \textit{SceneDiff}, achieve a higher FID and lower recognition accuracy compared to our methods and the text-conditioned baselines, \textit{MDM} and \textit{MLD}.
Since \textit{HUMANISE} and \textit{SceneDiff} are trained on limited synthetic 3D point clouds (e.g., 643 indoor scenes in ScanNet), these models struggle to generalize to real-world point clouds constructed from single images in diverse indoor and outdoor scenes, leading to lower motion quality.
Compared to the models conditioned on text alone, the advanced model \textit{MLD} achieves better metrics than \textit{MDM}.
By training on our large-scale human motion dataset with 300k sequences, \textit{MDM+} achieved a 72\% lower FID and a 90\% higher accuracy compared to \textit{MDM}, which is trained on the HumanML3D dataset with only 14k motion sequences. This result highlights the significant improvement in human motion generation enabled by training on our large-scale HiC-Motion dataset extracted from real-world videos.

Among models trained on the same backbone and dataset but with different input conditions (i.e., \textit{MDM+}, \textit{Ours-scene}, \textit{Ours}), \textit{Ours} achieves the lowest FID score, the highest accuracy and diversity. \textit{Ours} achieves 37\% higher accuracy compared to \textit{Ours-scene}, indicating that the in-context conditioning method effectively enables the model to generate actions aligned with specific prompts.
On the other hand, \textit{Ours-scene} achieves a higher multimodality score, which measures diversity within the same prompt. As \textit{Ours-scene} lacks the text constraint, it exhibits greater variation in outputs for identical prompts.

\begin{table}[t]
    \centering
    \scriptsize
    \caption{\textbf{Automated evaluation.} We report average VLM scores (0-5) for generated motions, assessing alignment with scene, text, and pose quality. Our method outperforms all evaluated baselines.}
        \tabcolsep=0.1cm
     \vspace{-1mm}
    \begin{tabular}{lcccc}
        \toprule
        Methods & Scene-Align~$(\uparrow)$ & Text-Align~$(\uparrow)$ & Quality~$(\uparrow)$ & Total~$(\uparrow)$ \\
        \midrule
        MDM~\cite{tevet2023human} & 2.25 & 1.35 & 1.50 & 5.10 \\
        MLD~\cite{chen2023executing} & 2.85 & 1.95 & 1.90 & 6.70 \\
        \midrule
        SceneDiff~\cite{huang2023diffusion} & 2.05 & 1.20 & 1.20 & 4.45\\
        HUMANISE~\cite{wang2022humanise} & 2.20 & 1.45 & 1.30 & 4.95 \\
        \midrule
        MDM+~\cite{tevet2023human} & 2.57 & 1.73 & 1.94 & 6.24 \\
        Ours-scene & 2.90 &  2.00 & 1.95 & 6.85 \\
        Ours & \bf 3.55 & \bf 2.70 & \bf 2.85 & \bf 9.10 \\
        \bottomrule
    \end{tabular}
    \label{tab:vlm}

\end{table}

\topic{Automated evaluation.} 
Since there is currently no established metric to assess the compatibility between generated motion sequences and 2D background images, we employ the vision-language model (VLM) ChatGPT-4o~\cite{openai2024chatgpt} for automated evaluation.
Given the generated SMPL pose rendered on the background image alongside the input text prompt, the VLM provides scores on a 0-5 scale for the following criteria: 1) alignment of the pose with the background, 2) alignment of the pose with the text prompt, and 3) overall quality of the generated pose.
For consistency, we use the middle frame of each generated sequence for evaluation. Average scores over 20 test set videos are reported in~\cref{tab:vlm}. 
Our method consistently outperforms the compared approaches across all criteria, achieving the highest score of 3.55 for alignment with the scene, demonstrating that our in-context framework effectively enforces affordance-aware motion generation on 2D scenes.

\begin{table}[t]
    \centering
    \scriptsize
    \caption{\textbf{Ablation study.} We study different transformer block designs, and choose \textit{AdaLN} for timestep conditioning and \textit{In-Context} for text and scene conditions as our main configuration. }
     \vspace{-1mm}
    \begin{tabular}{lll|cc}
        \toprule
        Timestep & Text & Scene & FID~$(\downarrow)$ & Accuracy~$(\uparrow)$ \\
        \midrule
        AdaLN & In-Context & In-Context & \bf 44.639 & \bf 0.661  \\
        AdaLN & In-Context & Cross-Attn & 47.656 & 0.567 \\
        In-Context & In-Context & In-Context & 62.927 & 0.554 \\
        In-Context & In-Context & Cross-Attn & 66.827 & 0.519\\
        \bottomrule
    \end{tabular}
    \label{tab:ablation}
    
\vspace{-1mm}
\end{table}

\topic{Ablation study.}
\label{sec:ablation}
As discussed in~\cref{sec:approach}, we train our 2D-conditioned motion diffusion models using various transformer block designs, including AdaLN, in-context, and cross-attention layers, to condition on the timestep, text, and scene inputs. We evaluate four models, each with a different combination of these conditioning methods in \cref{tab:ablation}.
Our results demonstrate that the model incorporating \textit{AdaLN} for timestep conditioning and \textit{In-Context} conditioning for text and scene inputs achieves the best FID and accuracy. Thus, we adopt this setup in our main framework.

\subsection{Motion-guided Human Video Generation}
One important downstream application supported by our approach is human video generation guided by motion sequence.
We employ a two-stage approach: first, given a scene image and text prompt, our model generates a scene-compatible motion sequence. Next, using this generated motion sequence and a reference human in the scene, we apply Champ~\cite{zhu2024champ} to animate the reference human guided by the generated motion, enabling the creation of affordance-aware human videos that align with the target background.
Additionally, we use Gen-3~\cite{gen1} to generate a motion-guided video. Although Gen-3 does not preserve the original background, our generated motion still serves as an effective guidance signal for the human subject, while the scene image provides the desired background’s layout and semantic information.
As shown in \cref{fig:champ}, the accurate and smooth motion sequences generated by our model allow both Champ and Gen-3 to produce videos with detailed human shapes and clean motion. Our method generates 256-frame sequences of complex activities such as dancing and playing tennis (see~\cref{fig:teaser}). 
We also include results from Stable Video Diffusion (SVD)~\cite{blattmann2023stablevideodiffusionscaling} using the same reference frame. Without pose guidance, SVD generates incomplete human geometry and inconsistent, blurry results, underscoring the advantages of using our method to generate intermediate pose sequences for video generation.

\section{Conclusions}
\label{sec:conclusion}
We introduced a novel task of generating human motion conditioned on a scene image. Our approach employs a conditional diffusion model enhanced by in-context learning techniques. To support this, we collected a large-scale dataset of diverse human activities and environments for model training. Our method effectively predicts 2D-aligned human motion and improves motion quality in video generation.
Despite these advancements, our framework does not control camera movement in generated motions, and the two-pass video generation pipeline has not been jointly optimized with our dataset. We leave these aspects for future work.

{
    \small
    \bibliographystyle{ieeenat_fullname}
    \bibliography{main}
}

\end{document}